\documentclass[10pt,twocolumn,letterpaper]{article}
\usepackage[pagenumbers]{cvpr}
\usepackage{multirow,colortbl}
\definecolor{cvprblue}{rgb}{0.21,0.49,0.74}
\usepackage[breaklinks,colorlinks,allcolors=cvprblue]{hyperref}
\hypersetup{
  pdftitle={Anchoring with Visual Evidence: Explicit Semantic Anchors for Joint Change Detection and Captioning},
  pdfauthor={Kai Zheng, Hang-Cheng Dong, Puyi Fang, Junjia Pan, Zhenkai Wu, Wei Zhang},
  pdfsubject={Joint remote sensing change detection and captioning},
  pdfkeywords={remote sensing, change detection, change captioning, explicit semantic anchors}
}
\DeclareMathOperator{\GAP}{GAP}
\DeclareMathOperator{\Pool}{Pool}
\DeclareMathOperator{\Resize}{Resize}
\DeclareMathOperator{\CE}{CE}
\DeclareMathOperator{\BCE}{BCE}

\DeclareMathOperator{\softmax}{softmax}
\DeclareMathOperator{\sg}{sg}
\DeclareMathOperator*{\argmax}{arg\,max}
\newcommand{\ind}{\mathbf{1}}

\providecolor{bestred}{RGB}{255,205,205}
\providecolor{secondblue}{RGB}{205,225,255}
\title{Anchoring with Visual Evidence: Explicit Semantic Anchors for Joint Change Detection and Captioning}
\author{%
Kai Zheng$^{1,\dagger}$\quad
Hang-Cheng Dong$^{2,3,\dagger}$\quad
Puyi Fang$^{4,\dagger}$\\[2pt]
Junjia Pan$^{1}$\quad
Zhenkai Wu$^{1}$\quad
Wei Zhang$^{1,*}$\\[5pt]
{\small $^{1}$Zhejiang University\quad $^{2}$Harbin Institute of Technology}\\
{\small $^{3}$Harbin Institute of Technology Suzhou Research Institute}\\
{\small $^{4}$Zhejiang Gongshang University}\\[3pt]
{\ttfamily\small zhengkai1990@zju.edu.cn\quad hunsen\_d@hit.edu.cn}\\
{\ttfamily\small cstzhangwei@zju.edu.cn}\\[3pt]
{\small $^{\dagger}$Equal contribution.\quad $^{*}$Corresponding author.}
}
\date{}

\begin{document}
\maketitle

\begin{abstract}
Joint change detection and captioning seeks to localize changes and
describe them from bi-temporal remote sensing images. A key challenge
is translating dense visual evidence into the semantic conditions
that guide description generation. We propose Explicit Semantic
Anchors (ESA), a supervised interface that exposes entity presence,
temporal appearance variation, and spatial extent. A prediction-guided
readout combines visual features and predicted change
probabilities, and three factor heads estimate the corresponding
anchor activations. These activations guide generation through two
complementary routes. The context route maps the three factor activations to three separate memory tokens, one per factor, and appends them to the dense visual memory, while the lexicon route produces an additive vocabulary bias at each decoding
step. Dense visual memory remains available for spatial detail and
information beyond the explicit factors. Anchor targets are derived
from existing images and change annotations, requiring no additional
anchor-level language annotation. At inference, all conditions are
computed from the input image pair and the model's own predictions.
Experiments on LEVIR-MCI and WHU-CDC show strong joint detection and
captioning performance, supporting the use of supervised visual
attributes as explicit conditions for change description generation.
\end{abstract}

\section{Introduction}
\label{sec:intro}

A change map and a change description offer complementary accounts of
how a scene evolves. In remote sensing, change detection (CD) localizes
changes at the pixel level, whereas change captioning (CC) expresses
their content in natural language. Multi-level change interpretation
(MCI) brings these tasks into a common framework~\cite{liu2024changeagent},
combining spatially precise predictions with descriptions that people
can readily interpret. Realizing this complementarity requires more
than producing two outputs from the same image pair: the description
should be informed by the visual evidence underlying the change map.
This motivates a central question: \emph{how can a joint model turn
dense change evidence into explicit semantic conditions for language
generation?}

The challenge lies in the different decisions required by the two
tasks. Detection assigns labels to spatial locations, while captioning
selects a compact set of attributes and organizes them into a sentence.
Even a category-aware change map does not uniquely determine a
description. A caption may need to identify the affected entities,
interpret temporal appearance differences, and distinguish a localized
alteration from a more extensive transformation. Dense bi-temporal
features contain evidence for these decisions, but their presence in a
decoder's visual memory does not expose which attributes have been
inferred or how they influence word selection. An explicit intermediate
representation would make part of this mapping available for direct
supervision and inspection.

Recent work has strengthened the connection between visual perception
and language through shared representations and task interaction.
Joint models such as Change-Agent and ChangeMinds support detection and
captioning within a common architecture~\cite{liu2024changeagent,wang2024changeminds}.
LCCC introduces critique and correction between the two
tasks~\cite{li2026lccc}, while CoDeRNet selectively transfers features
according to task-specific support and reliability
cues~\cite{cho2026codernet}. In change captioning, HiSem explores
hierarchical semantic organization~\cite{wang2026hisem}, and STAND
introduces semantic anchoring constraints~\cite{gong2026stand}.
These approaches motivate a complementary question about the content
of semantic guidance: \emph{which attributes can be grounded in the
available visual supervision, and where should their predictions enter
the captioning process?} We address this question through a supervised
interface between predicted change regions and language generation.

Our starting point is that the visual learning problem already
provides useful supervision for such an interface. Change masks
identify affected regions and, when semantic labels are available,
their categories. The ordered image pair provides appearance
differences, and mask statistics quantify the fraction of the scene
that has changed. Masks Can Talk demonstrates the value of deriving
structured textual supervision from existing change
annotations~\cite{zheng2026maskscantalk}. We use this available structure
to supervise intermediate visual attributes that can subsequently
condition a caption decoder. This reuses the existing images and masks
without requiring additional anchor-level language annotations.

We propose \textbf{Explicit Semantic Anchors (ESA)}, an interface that
reads predicted change evidence and exposes three complementary
factors: \emph{entity presence}, \emph{temporal appearance variation},
and \emph{spatial extent}. A shared visual pathway produces multi-scale
bi-temporal features and change probabilities. A prediction-guided
readout then combines global differences, category-conditioned
features, ordered temporal evidence, and region statistics. Three
supervised heads map this evidence to named factor states. Entity
states indicate which supported categories are present; temporal
states provide a weak appearance-polarity cue; and extent states
describe changed-area coverage. Their definitions make the predicted
conditions inspectable before decoding. The temporal factor is
supervised by intensity variation within changed regions, so it serves
as an appearance cue rather than a direct label for events such as
construction or demolition.

ESA reuses the same factor activations through a \textbf{context route}
and a \textbf{lexicon route}. The context route converts the three factor activations into three separate memory tokens and appends them to the
dense visual memory. This allows the decoder to attend to
the predicted attributes while constructing a sentence. The lexicon
route maps the same activations to an additive vocabulary bias,
providing an image-dependent influence on lexical scores at every
decoding step. The two routes therefore expose the same evidence at
complementary points in generation: contextual decoding and word
selection. Dense visual memory remains available for spatial detail
and content beyond the three factors. Standard caption supervision
learns how these conditions should inform the generated description.

During training, auxiliary targets are derived from the existing
visual annotations and image attributes, whereas the anchor readout
always receives features and predicted change probabilities. At
inference, the image pair alone supplies the change map, anchor
activations, and autoregressively generated caption. Experiments on
LEVIR-MCI and WHU-CDC evaluate the resulting model through detection
accuracy, captioning metrics, and qualitative comparisons. They show
strong joint performance, with the relative advantages varying across
datasets and metrics.

Our contributions are threefold:
\begin{itemize}
    \item We introduce an explicit semantic interface for MCI that
    organizes prediction-guided visual evidence into supervised entity,
    temporal-appearance, and spatial-extent factors, making these
    intermediate conditions available for inspection.
    \item We design a context route and a lexicon route that reuse the same anchor activations as three separate memory tokens and an additive vocabulary bias, connecting visual attributes to both contextual decoding and lexical selection.
    \item We derive anchor supervision from existing images and change
    annotations, enabling image-only inference without additional
    anchor-level language annotation, and evaluate the framework on
    two joint detection--captioning benchmarks.
\end{itemize}

\section{Related work}
\label{sec:related work}

\subsection{Remote Sensing Change Captioning}
Remote sensing change captioning (CC) translates changes between bi-temporal images into natural-language descriptions.
A major line of research improves the encoding of temporal differences for language generation.
RSICCformer~\cite{liu2022rsiccformer} introduced the LEVIR-CC benchmark and a dual-branch Transformer, while Chg2Cap~\cite{chang2023chg2cap} combines hierarchical attention with difference-aware feature refinement.
Beyond visual difference encoding, PromptCC~\cite{liu2023promptcc} separates change-existence classification from caption generation and connects visual representations to a pretrained language model through multiple prompts.
HiSem~\cite{wang2026hisem} further organizes change semantics through discrepancy-aware interaction, image-level routing, and token-level mixture-of-experts modeling.

Another line strengthens the spatial or semantic evidence used for description.
MaskApproxNet~\cite{sun2025maskapproxnet} estimates change masks through conditional diffusion and applies frequency-guided feature fusion before autoregressive caption decoding.
For building-damage captioning, KGBDCNet~\cite{wang2026kgbdcnet} combines keyword-guided cross-modal retrieval with spatiotemporal feature modeling to emphasize damage-related content.
These approaches illustrate how change recognition, spatial cues, and semantic guidance support caption generation.
CC evaluates change understanding through language. Joint MCI additionally requires explicit spatial predictions and evaluation of both outputs.

\subsection{Remote Sensing Change Detection}
Remote sensing change detection (CD) predicts spatially explicit changes from multi-temporal imagery.
Early fully convolutional Siamese networks~\cite{daudt2018fully} fuse features from shared-weight temporal encoders through concatenation or absolute differences in skip connections.
To incorporate broader context, BIT~\cite{chen2022bit} models interactions among compact semantic tokens and projects their context back to dense image features.
ChangeFormer~\cite{bandara2022changeformer} instead combines a hierarchical Siamese Transformer encoder with an MLP decoder for multiscale change prediction.
Extending this architectural progression, ChangeMamba~\cite{chen2024changemamba} introduces spatiotemporal state-space modeling for binary and semantic change detection as well as building damage assessment.

Recent work also expands the semantic scope of detection.
OpenDPR~\cite{guo2026opendpr} addresses open-vocabulary CD by combining foundation-model change proposals with diffusion-guided visual prototype retrieval for category identification.
Thus, CD encompasses both change localization and structured semantic prediction, including change categories or damage levels.
These structured outputs complement the natural-language descriptions targeted by CC.

\subsection{Multi-Level Change Interpretation}

Multi-level change interpretation (MCI) unifies CD and CC to jointly predict pixel-level change maps and semantic descriptions. Change-Agent~\cite{liu2024changeagent} introduced LEVIR-MCI and a joint perception model with bi-temporal iterative interaction, while CTMTNet~\cite{shi2024ctmtnet} provided the LEVIR-CDC and WHU-CDC datasets through attention-based enhancement and bi-temporal fusion. Building on these formulations, ChangeMinds~\cite{wang2024changeminds} uses ChangeLSTM and a unified decoder to serve both heads, and Semantic-CC~\cite{zhu2024semanticcc} employs a SAM-based encoder with pixel-level guidance coordinated through a three-stage training strategy, connecting language generation to spatially grounded representations. Recent methods further address this coupling with distinct mechanisms. Change-UP~\cite{yang2025changeup} integrates adaptive representations and text-driven style diversification, LCCC~\cite{li2026lccc} couples detection-derived change tokens with a critique--correction framework, and CoDeRNet~\cite{cho2026codernet} selectively routes features based on receiver-side support needs and sender-side reliability cues. A central issue across these designs is how to exchange complementary task evidence while preserving the spatial detail required by CD and the semantic selectivity required by CC.

\section{Method}
\label{sec:method}

\subsection{Problem Formulation and Overview}
\label{sec:overview}
\begin{figure*}[t]
\centering
\includegraphics[width=\textwidth]{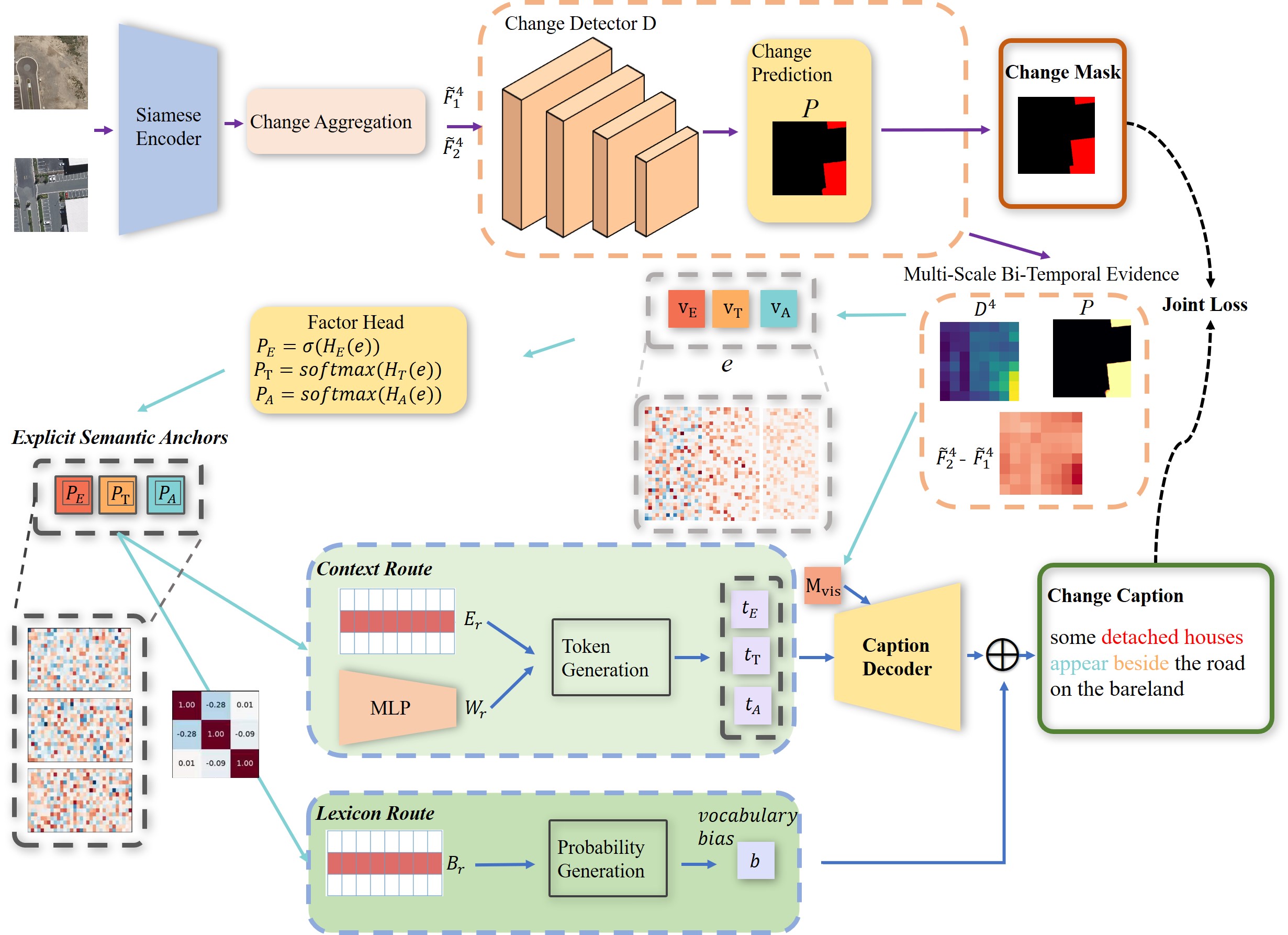}
\caption{Overview of Explicit Semantic Anchors (ESA). A shared visual pathway
predicts the change map and supplies dense visual memory. A prediction-guided,
stop-gradient readout exposes change evidence to three factor heads for entity
presence, temporal appearance variation, and spatial extent. The context route
maps the three factor activations to three factor-specific tokens and appends
them, separately and without aggregation, to the dense visual memory;
the decoder can attend to each factor independently. The lexicon route maps the
same activations to an additive bias on the output vocabulary logits before the
softmax. Dashed arrows indicate auxiliary supervision derived from
existing masks and image intensities during training; inference requires only
the image pair.}
\label{fig:method}
\end{figure*}

Given co-registered images $X_1,X_2\in\mathbb{R}^{3\times H\times W}$, MCI predicts a change map $\widehat{Y}$ and a caption $w=(w_1,\ldots,w_L)$. The detection ontology contains $C$ classes, including background indexed by $0$. Training provides a change map $Y$ and reference captions $\mathcal{R}$. Figure~\ref{fig:method} summarizes the framework. A shared visual pathway supports both outputs, while the ESA pathway reads the predicted visual evidence and supplies explicit conditions to the caption decoder. Ground-truth annotations construct auxiliary targets during training; the forward-pass anchor inputs always use model features and predictions.

ESA exposes three interpretable factors: entity presence, temporal appearance variation, and spatial extent. These factors summarize complementary attributes that can inform a change description. Their auxiliary targets are derived from the existing masks and ordered images, without parsing captions or collecting additional anchor-level language annotations. The resulting activations condition caption generation through a \emph{context route} and a \emph{lexicon route}, while dense visual memory remains available for details beyond these global factors.

\subsection{Multi-scale Bi-temporal Evidence}
\label{sec:visual}
\paragraph{Shared encoding.}
A hierarchical backbone $E$ extracts four feature stages at strides $4$, $8$, $16$, and $32$. A stage-specific $1\times1$ convolution $\pi_s$ projects each stage to a common channel width $d$, followed by layer normalization:
\begin{equation}
F_i^s=\mathrm{LN}\!\left(\pi_s(E_s(X_i))\right),
\label{eq:shared_encoding}
\end{equation}
where $\quad i\in\{1,2\}$ and $\ s\in\{1,2,3,4\}$. The backbone and projection parameters are shared between observations; $s=4$ denotes the coarsest grid.

\paragraph{Change aggregation and enhancement.}
We retain both absolute and ordered differences, $A^s=|F_1^s-F_2^s|$ and $R^s=F_1^s-F_2^s$. A convolutional gate modulates a residual on the later observation:
\begin{align}
G^s&=\sigma\!\left(\phi_g^s([F_1^s;F_2^s;A^s;R^s])\right),
\label{eq:gate}\\
\widetilde{F}_1^s&=F_1^s,\qquad
\widetilde{F}_2^s=F_2^s+\phi_a^s(A^s)\odot G^s.
\label{eq:enhancement}
\end{align}
Here $[\,;\,]$ denotes channel concatenation, $\phi_g^s$ is a $1\times1$ convolution, and $\phi_a^s$ is a $3\times3$ convolution block with batch normalization and ReLU. The earlier observation provides a reference, while the later feature receives difference-conditioned information. The learned residual is not constrained to increase the distance between the two features. Multi-scale change descriptors are also aggregated through a top-down recursion:
\begin{equation}
\begin{aligned}
C^4&=\phi_c^4(A^4),\\
C^s&=\phi_c^s(A^s)+\mathrm{Up}\!\left(\eta_s(C^{s+1})\right),
\end{aligned}
\label{eq:aggregation}
\end{equation}
for $s=3,2,1$, where $\eta_s$ is a $1\times1$ convolution that matches the channel width before upsampling.

\paragraph{Dense change prediction.}
At each scale, a convolutional block encodes the ordered enhanced pair and its differences:
\begin{equation}
\begin{aligned}
D^s=\phi_d^s\big([&\widetilde{F}_1^s;\widetilde{F}_2^s;
|\widetilde{F}_1^s-\widetilde{F}_2^s|;\widetilde{F}_1^s-\widetilde{F}_2^s]\big).
\end{aligned}
\label{eq:dense_features}
\end{equation}
A top-down decoder fuses these features with $Q^4=D^4$ and $Q^s=\psi_s(D^s+\mathrm{Up}(Q^{s+1}))$. A $1\times1$ classifier followed by resizing produces logits $Z_{\mathrm{cd}}\in\mathbb{R}^{C\times H\times W}$. The probability map and final prediction are
\begin{equation}
\begin{aligned}
P &= \softmax(Z_{\mathrm{cd}}),\\
\widehat{Y}(u) &= \argmax_{c\in\{0,\ldots,C-1\}} P_c(u).
\end{aligned}
\label{eq:detection}
\end{equation}
The maximization includes background. Ordered features preserve temporal information that absolute differences alone discard.

\subsection{Explicit Semantic Anchors}
\label{sec:anchors}
\paragraph{Prediction-guided evidence readout.}
The anchor readout combines global differences, category-conditioned content, ordered temporal evidence, and change coverage. We resize $P$ to the coarsest grid, obtaining $\bar{P}_c=\Resize_{H_4,W_4}(P_c)$ and $\bar{P}_{\mathrm{fg}}=\sum_{c=1}^{C-1}\bar{P}_c$. Normalized weighted pooling is defined as
\begin{equation}
\Pool(Z,q)=\frac{\sum_u q(u)Z(u)}{\sum_u q(u)+\epsilon},
\label{eq:pooling}
\end{equation}
where $\epsilon>0$ stabilizes the denominator. Since pooling normalizes by the total mask weight, area statistics are supplied separately. The evidence components are
\begin{align}
v_g&=\GAP(D^4),\quad
v_c=\Pool(D^4,\bar{P}_c),
\label{eq:entity_evidence}\\
v_T&=\Pool(\widetilde{F}_2^4-\widetilde{F}_1^4,\bar{P}_{\mathrm{fg}}),
\label{eq:temporal_evidence}\\
a_c&=\frac{1}{HW}\sum_u P_c(u),\quad
a_{\mathrm{fg}}=\sum_{c=1}^{C-1}a_c.
\label{eq:area_evidence}
\end{align}
The index $c$ in Eqs.~\eqref{eq:entity_evidence} and~\eqref{eq:area_evidence} ranges over foreground categories. We concatenate the components and detach their visual inputs:
\begin{equation}
\begin{aligned}
e=\sg\big([&v_g;v_1;\ldots;v_{C-1};v_T;\\[-1mm]
&a_1;\ldots;a_{C-1};a_{\mathrm{fg}}]\big).
\end{aligned}
\label{eq:evidence}
\end{equation}
For descriptor width $d$, the vector has dimension $(C+1)d+C$. The temporal readout uses later-minus-earlier features, whereas $R^s$ uses the opposite sign convention; both are fixed as defined above. The stop-gradient operation blocks updates to the visual inputs through this auxiliary evidence path. Other paths through the shared visual representation remain trainable.

\paragraph{Factor activations.}
Three heads $H_r$, each comprising layer normalization, a linear layer, GELU, and a final linear layer, predict the factor activations:
\begin{equation}
\begin{aligned}
p_E&=\sigma(H_E(e))\in[0,1]^C,\\
p_T&=\softmax(H_T(e))\in[0,1]^3,\\
p_A&=\softmax(H_A(e))\in[0,1]^3.
\end{aligned}
\label{eq:factor_heads}
\end{equation}
The entity head contains independent indicators for the $C-1$ foreground categories and a separate no-change indicator. The temporal head encodes neutral, positive, and negative appearance variation; the extent head represents small, medium, and large changed-area fractions. These are global factors, rather than separate event assignments for individual objects. Their named states and auxiliary targets make the predictions inspectable, while learnable readout tables translate them into generation conditions. Entity activations are independent Bernoulli probabilities, so their weighted readout need not be a convex combination.

\subsection{Reusing Existing Visual Supervision}
\label{sec:anchor_supervision}
Auxiliary targets are constructed from the existing visual annotations and image attributes. They require neither a reference-caption parser nor externally generated textual descriptors.

\paragraph{Entity targets.}
Foreground presence and the no-change target follow from the detection map:
\begin{equation}
\begin{aligned}
y_{E,c}&=\ind[\exists u:Y(u)=c],\quad c=1,\ldots,C-1,\\
y_{E,\varnothing}&=\ind[\forall u:Y(u)=0].
\end{aligned}
\label{eq:entity_targets}
\end{equation}
The supported entity vocabulary follows the mask ontology. Road and building annotations in LEVIR-MCI support those category labels, whereas a binary mask alone does not supply arbitrary semantic categories.

\paragraph{Temporal-variation targets.}
Let $M(u)=\ind[Y(u)>0]$, and let $\ell(X_i)(u)$ denote the scalar image-intensity representation. We compute the mean intensity difference over changed regions:
\begin{equation}
\delta_\ell=\frac{\sum_u M(u)[\ell(X_2)(u)-\ell(X_1)(u)]}{\sum_u M(u)+\epsilon},
\label{eq:intensity_delta}
\end{equation}
and assign the three-state target
\begin{equation}
y_T=\begin{cases}
1,&\delta_\ell>\tau_T,\\
2,&\delta_\ell<-\tau_T,\\
0,&\text{otherwise}.
\end{cases}
\label{eq:temporal_targets}
\end{equation}
This target provides a weak photometric-polarity cue. The neutral state includes both unchanged samples and changes with a small mean intensity difference. These labels do not directly supervise construction or demolition; the association between appearance evidence and captioned events is learned through caption supervision.

\paragraph{Extent targets.}
For changed-area fraction $r=(HW)^{-1}\sum_u M(u)$, we define
\begin{equation}
y_A=\begin{cases}
0,&r\leq\tau_s,\\
1,&\tau_s<r\leq\tau_l,\\
2,&r>\tau_l.
\end{cases}
\label{eq:extent_targets}
\end{equation}
The bins describe spatial coverage, rather than object count or individual object size. The thresholds are dataset-specific hyperparameters selected using training data.

\subsection{Anchor-conditioned Caption Generation}
\label{sec:routes}
For each factor $r\in\{E,T,A\}$, we learn an embedding table $E_r\in\mathbb{R}^{K_r\times d}$ and a vocabulary table $B_r\in\mathbb{R}^{K_r\times|\mathcal{V}|}$, where $(K_E,K_T,K_A)=(C,3,3)$ and $\mathcal{V}$ is the caption vocabulary. Both routes read the same image-dependent factor activations.

\paragraph{Context route.}
The embedding tables map the three factor activations to
three factor-specific tokens:
\begin{equation}
t_r = p_r^{\top} E_r, \qquad r \in \{E, T, A\}.
\end{equation}
Let $M_{\mathrm{vis}} = [m_1,\dots,m_N] \in \mathbb{R}^{N\times d}$ denote the
dense visual memory of the caption decoder, inherited from the base captioner
\cite{liu2024changeagent}: it consists of the flattened features
$\{X^{s}\}_{s=1}^{4}$ at the four scales ($H_sW_s$ tokens at scale $s$) followed
by the mask-conditioned global token $g$, so that
$N = 1 + \sum_{s=1}^{4} H_sW_s$. The decoder cross-attends over
$M = [M_{\mathrm{vis}}; t_E; t_T; t_A]$; the three anchor tokens therefore
extend the memory from $N$ to $N+3$ entries instead of replacing the spatial
evidence.

The global token is
\begin{equation}
g = \sum_{s=1}^{4} \mathrm{GAP}\!\left(\tilde F_2^{s} \odot
   \mathrm{sg}(\bar P_{\mathrm{fg}}^{s})\right) + \mathrm{GAP}(A^{4}),
\label{eq:anchor_g}
\end{equation}
where $\bar P_{\mathrm{fg}}^{s}$ is the predicted foreground probability
resized to the grid of stage $s$. The three tokens are kept separate; they are
neither summed nor averaged. The context route appends them, as three
additional memory entries, to the dense visual memory:
\begin{equation}
M = [M_{\mathrm{vis}}; t_E; t_T; t_A].
\label{eq:memory}
\end{equation}

The decoder can therefore attend to the entity, temporal, and extent conditions
independently. This three-token construction is used in all main experiments.

\paragraph{Lexicon route.}
The vocabulary tables map the same factors to an additive bias over candidate words:
\begin{equation}
b_{\mathrm{anc}}=p_E^{\top}B_E+p_T^{\top}B_T+p_A^{\top}B_A.
\label{eq:lexicon_bias}
\end{equation}
The tables are initialized to zero. The bias is added to the decoder logits at each autoregressive step:
\begin{equation}
\begin{aligned}
&p(w_t\mid w_{<t},X_1,X_2)\\
&\qquad=\softmax\!\left(W_o h_t+\lambda_b b_{\mathrm{anc}}\right).
\end{aligned}
\label{eq:lexicon_route}
\end{equation}
The context route conditions the decoder memory, and the lexicon route directly conditions lexical scores. The readout tables are shared model parameters, while their activations and outputs vary with the input pair. Since dense visual memory remains accessible, the anchors provide an explicit conditioning interface rather than a strict information bottleneck.

\subsection{Optimization and Inference}
\label{sec:optimization}
Entity supervision uses binary cross-entropy, while temporal variation and spatial extent use categorical cross-entropy:
\begin{equation}
\begin{aligned}
\mathcal{L}_{\mathrm{anc}}={}&\lambda_E\BCE(H_E(e),y_E)\\
&+\lambda_T\CE(H_T(e),y_T)\\
&+\lambda_A\CE(H_A(e),y_A).
\end{aligned}
\label{eq:anchor_loss}
\end{equation}
The loss arguments are logits. For multiple reference captions, the caption objective is
\begin{equation}
\begin{aligned}
\mathcal{L}_{\mathrm{CC}}={}&-\frac{1}{|\mathcal{R}|}\sum_{w\in\mathcal{R}}\frac{1}{|w|}\\
&\times\sum_{t=1}^{|w|}\log p(w_t\mid w_{<t},X_1,X_2).
\end{aligned}
\label{eq:caption_loss}
\end{equation}
If label smoothing is used, soft token targets replace the token-level negative log likelihood. Its coefficient is distinct from the intensity threshold $\tau_T$. The complete objective is
\begin{equation}
\begin{aligned}
\mathcal{L}&=\mathcal{L}_{\mathrm{MCI}}+\mathcal{L}_{\mathrm{anc}},\\
\mathcal{L}_{\mathrm{MCI}}&=\lambda_{\mathrm{CD}}(\mathcal{L}_{\mathrm{CE}}+
\lambda_{\mathrm{Dice}}\mathcal{L}_{\mathrm{Dice}})+\lambda_{\mathrm{CC}}\mathcal{L}_{\mathrm{CC}},
\end{aligned}
\label{eq:overall_loss}
\end{equation}
where $\mathcal{L}_{\mathrm{CE}}$ and $\mathcal{L}_{\mathrm{Dice}}$ supervise dense change prediction.

The auxiliary losses update the anchor heads but do not reach the visual inputs through Eq.~\eqref{eq:evidence}. Caption gradients train the heads and readout tables through both routes, and can also update the shared visual pathway through $M_{\mathrm{vis}}$. Detaching the anchor evidence therefore does not freeze the detector. At inference, the model computes $P$, predicts the three factor activations, and generates the caption autoregressively. Ground-truth masks, auxiliary targets, and reference captions are absent. The images are encoded once, followed by the decoding steps needed to generate the sentence.

\section{Experiments}
\label{sec:experiments}

\subsection{Datasets}
\label{sec:datasets}
We evaluate on LEVIR-MCI~\cite{liu2024changeagent} and WHU-CDC~\cite{shi2024ctmtnet}, which provide bi-temporal image pairs, pixel-level change masks, and five reference captions per pair. Both use $256\times256$ images. LEVIR-MCI uses 6,815/1,333/1,929 pairs for training/validation/testing.

\subsection{Evaluation Metrics}
\label{sec:metrics}
We report mIoU for detection and BLEU-1/2/3/4~\cite{papineni2002bleu}, METEOR~\cite{banerjee2005meteor}, ROUGE-L~\cite{lin2004rouge}, and CIDEr-D~\cite{vedantam2015cider} for captioning. The caption summary $S_m^*$ averages BLEU-4, METEOR, ROUGE-L, and CIDEr-D; the joint score TS averages mIoU and $S_m^*$. All metrics are higher-is-better.

\subsection{Quantitative Comparison}
\label{sec:quantitative}

The compared methods are grouped into detection-only, captioning-only, and joint CD--CC approaches. The first two groups provide task-specific reference points, while the joint methods permit direct inspection of both outputs.

\begin{table*}[htpb]
\centering
\caption{Comprehensive performance comparison on LEVIR-MCI. Detection-only methods report mIoU, captioning-only methods report text metrics, and joint CD-CC methods report both. All metrics are higher-is-better. Best and second-best results are highlighted with red and blue backgrounds, respectively.}
\label{tab:main_joint_merged}
\definecolor{bestred}{RGB}{255,205,205}
\definecolor{secondblue}{RGB}{205,225,255}
\renewcommand{\arraystretch}{1.15}
\setlength{\tabcolsep}{3.0pt}
\resizebox{\textwidth}{!}{
\begin{tabular}{c|c|c|ccccccc|c|c}
\noalign{\hrule height 1.2pt}
Tasks & Methods & mIoU & BLEU-1 & BLEU-2 & BLEU-3 & BLEU-4 & METEOR & ROUGE-L & CIDEr-D & \(S_m^*\) &  TS \\
\hline
\multirow{6}{*}{\begin{tabular}[c]{@{}c@{}}Change\\Detection\end{tabular}}
& BIT & 85.10 & -- & -- & -- & -- & -- & -- & -- & -- & -- \\
& CDxLSTM & 86.02 & -- & -- & -- & -- & -- & -- & -- & -- & -- \\
& ChangeFormer & 85.26 & -- & -- & -- & -- & -- & -- & -- & -- & -- \\
& ChangeMamba & 86.16 & -- & -- & -- & -- & -- & -- & -- & -- & -- \\
& CDMamba & 84.48 & -- & -- & -- & -- & -- & -- & -- & -- & -- \\
& ChangeDA & 85.64 & -- & -- & -- & -- & -- & -- & -- & -- & -- \\
\hline
\multirow{5}{*}{\begin{tabular}[c]{@{}c@{}}Change\\Captioning\end{tabular}}
& Chg2Cap & -- & 82.08 & 74.01 & 67.22 & 61.93 & 37.94 & 72.63 & 127.82 & 75.08 & -- \\
& RSCaMa & -- & 85.83 & 77.80 & 70.66 & 64.72 & 40.16 & 75.06 & 136.40 & 79.09 & -- \\
& PromptCC & -- & 83.66 & 75.73 & 69.10 & 63.54 & 38.82 & 73.72 & 136.44 & 78.13 & -- \\
& Pix4Cap & -- & 85.56 & 77.08 & 69.79 & 63.78 & 39.96 & 75.12 & 136.76 & 78.91 & -- \\
& CCExpert & -- & 86.65 & 78.47 & 71.31 & 65.49 & \cellcolor{secondblue}41.82 & 76.55 & 143.32 & 81.80 & -- \\
\hline
\multirow{7}{*}{\begin{tabular}[c]{@{}c@{}}Change Detection\\\& Captioning\end{tabular}}
& MCINet & 86.28 & 86.33 & 78.47 & 71.42 & 65.48 & 40.91 & 75.87 & 140.53 & 80.70 & 83.49 \\
& ChangeMinds & 83.25 & 83.58 & 72.28 & 63.33 & 55.96 & 39.22 & \cellcolor{bestred}77.18 & 133.26 & 76.41 & 79.83 \\
& FST-Net & 85.32 & 86.76 & 78.82 & 71.71 & 65.67 & 40.51 & 76.15 & 140.04 & 80.59 & 82.96 \\
& Change-UP & 87.02 & 86.87 & 79.10 & 71.98 & 66.20 & 41.80 & \cellcolor{secondblue}76.61 & \cellcolor{secondblue}143.40 & \cellcolor{secondblue}82.00 & 84.51 \\
& LCCC & 86.50 & 87.07 & 79.09 & 72.04 & 66.19 & 41.10 & 76.58 & 142.73 & 81.65 & 84.08 \\
& CoDeRNet & \cellcolor{bestred}87.79 & \cellcolor{secondblue}87.25 & \cellcolor{secondblue}79.42 & \cellcolor{secondblue}72.26 & \cellcolor{secondblue}66.27 & 41.08 & 76.36 & 141.79 & 81.38 & \cellcolor{secondblue}84.59 \\
& \textbf{Ours} & \cellcolor{secondblue}87.38 & \cellcolor{bestred}89.50 & \cellcolor{bestred}81.70 & \cellcolor{bestred}75.57 & \cellcolor{bestred}70.57 & \cellcolor{bestred}43.10 & 74.60 & \cellcolor{bestred}145.05 & \cellcolor{bestred}83.33 & \cellcolor{bestred}85.36 \\
\noalign{\hrule height 1.2pt}
\end{tabular}}
\end{table*}

\providecolor{bestred}{RGB}{255,205,205}
\providecolor{secondblue}{RGB}{205,225,255}

\begin{table*}[htpb!]
\centering
\caption{Comprehensive performance comparison on WHU-CDC. Detection-only methods report mIoU, captioning-only methods report text metrics, and joint CD-CC methods report both. All metrics are higher-is-better. Best and second-best results are highlighted with red and blue backgrounds, respectively.}
\label{tab:whucd_joint_merged}
\renewcommand{\arraystretch}{1.15}
\setlength{\tabcolsep}{3.0pt}
\resizebox{\textwidth}{!}{
\begin{tabular}{c|c|c|ccccccc|c|c}
\noalign{\hrule height 1.2pt}
Tasks & Methods & mIoU & BLEU-1 & BLEU-2 & BLEU-3 & BLEU-4 & METEOR & ROUGE-L & CIDEr-D & \(S_m^*\) & TS\\
\hline
\multirow{4}{*}{\begin{tabular}[c]{@{}c@{}}Change\\Detection\end{tabular}}
& BIT & 89.11 & -- & -- & -- & -- & -- & -- & -- & -- & -- \\
& CDXLSTM & 86.66 & -- & -- & -- & -- & -- & -- & -- & -- & -- \\
& ChangeFormer & 90.29 & -- & -- & -- & -- & -- & -- & -- & -- & -- \\
& ChangeMamba & 90.12 & -- & -- & -- & -- & -- & -- & -- & -- & -- \\
\hline
\multirow{2}{*}{\begin{tabular}[c]{@{}c@{}}Change\\Captioning\end{tabular}}
& RSICCFormer & -- & 80.05 & 74.24 & 69.61 & 66.54 & 42.65 & 73.91 & 133.44 & 79.14 & -- \\
& PromptCC & -- & 82.68 & 76.94 & 72.20 & 68.99 & 43.78 & 76.14 & 137.10 & 81.50 & -- \\
\hline
\multirow{6}{*}{\begin{tabular}[c]{@{}c@{}}Change Detection\\\& Captioning\end{tabular}}
& MCINet & 91.36 & 80.97 & 74.54 & 68.85 & 63.69 & 43.50 & 78.37 & 148.94 & 83.63 & 87.50 \\
& ChangeMinds & 89.55 & 81.09 & 77.24 & 74.89 & 73.48 & 46.13 & \cellcolor{secondblue}83.41 & 144.40 & 86.86 & 88.21 \\
& FST-Net & 87.36 & 88.15 & 83.46 & 79.55 & \cellcolor{secondblue}76.78 & 48.78 & 82.91 & 160.01 & 92.12 & 89.74 \\
& LCCC & 93.83 & \cellcolor{secondblue}89.03 & \cellcolor{secondblue}84.26 & \cellcolor{bestred}80.35 & \cellcolor{bestred}77.29 & \cellcolor{bestred}50.03 & \cellcolor{bestred}83.86 & \cellcolor{secondblue}164.99 & \cellcolor{bestred}94.04 & \cellcolor{secondblue}93.94 \\
& CoDeRNet & \cellcolor{secondblue}94.26 & 87.31 & 82.41 & 78.35 & 75.15 & 48.90 & 83.00 & 161.32 & 92.09 & 93.18 \\
& \textbf{Ours} & \cellcolor{bestred}94.86 & \cellcolor{bestred}89.99 & \cellcolor{bestred}84.69 & \cellcolor{secondblue}80.22 & 76.57 & \cellcolor{secondblue}48.95 & 79.27 & \cellcolor{bestred}169.97 & \cellcolor{secondblue}93.69 & \cellcolor{bestred}94.28 \\
\noalign{\hrule height 1.2pt}
\end{tabular}}
\end{table*}

\paragraph{Results on LEVIR-MCI.}
Table~\ref{tab:main_joint_merged} shows that ESA achieves the highest BLEU-1/2/3/4, METEOR, and CIDEr-D among the compared methods. Its BLEU-4 of 70.57 exceeds the strongest competing value, 66.27 from CoDeRNet, by 4.30 points. The BLEU-1, BLEU-2, and BLEU-3 improvements over CoDeRNet are 2.25, 2.28, and 3.31 points, respectively. ESA also improves CIDEr-D from 143.40 for Change-UP to 145.05 and raises the caption summary from 82.00 to 83.33. These results indicate stronger reference agreement across several complementary captioning measures.

For detection, ESA obtains 87.38 mIoU, the second-highest value in the table. It improves on MCINet, Change-UP, and LCCC by 1.10, 0.36, and 0.88 points, respectively, while remaining 0.41 points below CoDeRNet. The resulting TS is 85.36, compared with 84.59 for CoDeRNet. The improvements are not uniform: ESA's ROUGE-L of 74.60 is below the best result of 77.18 from ChangeMinds. Thus, the principal advantage on LEVIR-MCI is a stronger overall captioning profile with competitive localization, rather than dominance of every detection and language metric.

\begin{figure*}[t]
\centering
\includegraphics[width=\textwidth]{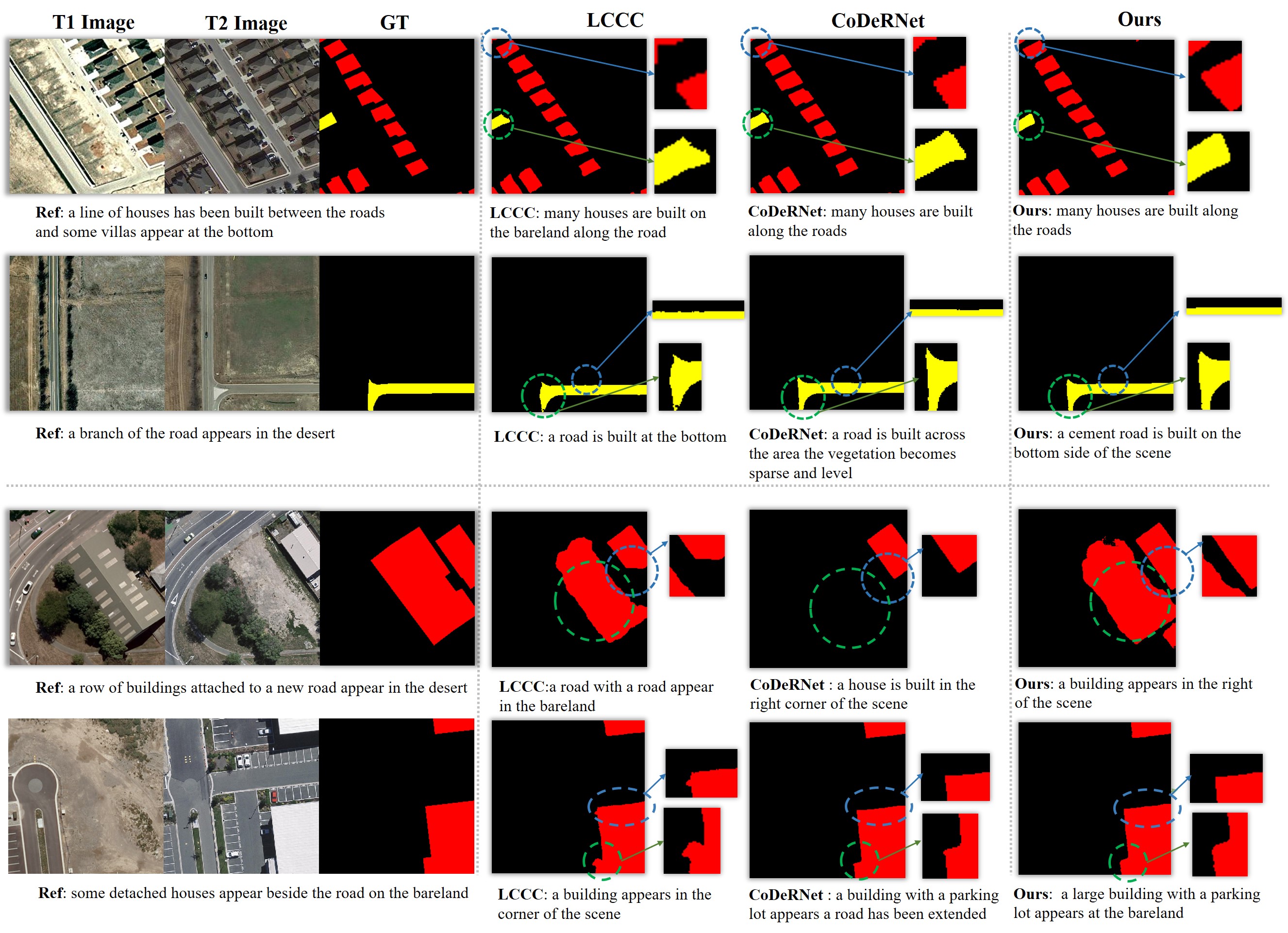}
\caption{Visual comparison of joint change detection and captioning results. The top two rows are from LEVIR-MCI, and the bottom two rows are from WHU-CDC. From left to right: pre-change (T1), post-change (T2), ground truth (GT), LCCC, CoDeRNet and Ours. In GT, red and yellow denote building and road changes, respectively. Dashed boxes highlight boundary details.}
\label{fig:cd_cc_visual}
\end{figure*}

\paragraph{Results on WHU-CDC.}
Table~\ref{tab:whucd_joint_merged} shows that ESA achieves 94.86 mIoU and 169.97 CIDEr-D on WHU-CDC, the highest values among the compared methods.

\subsection{Qualitative Comparison}
\label{sec:qualitative}

Figure~\ref{fig:cd_cc_visual} compares predicted maps and generated descriptions with LCCC and CoDeRNet. In the first example, ESA describes the group of houses built along the roads, capturing the principal changed entity and its spatial arrangement. This description is similar to CoDeRNet's and does not include every detail in the reference, such as the additional villas at the bottom. The corresponding detection map identifies the repeated building footprints and the road-change region. In the second example, ESA focuses on the newly appearing road at the bottom of the scene, while the magnified crops permit a closer comparison of the thin road segment and its junction. These cases illustrate how the map and sentence provide complementary accounts of the same change.

The building examples expose differences in foreground coverage and boundary shape. In the last example, ESA describes a large building together with a parking lot, and the predicted map delineates the main building-change regions. Such examples support inspection of entity selection, spatial wording, and localization together; they do not establish that every generated attribute is correct. In particular, the global anchor factors do not explicitly assign an event or a count to each object, and the dense visual memory remains necessary for finer spatial and contextual details.

\section{Conclusion}
\label{sec:conclusion}

We presented Explicit Semantic Anchors (ESA) for joint remote sensing change detection and captioning. ESA reads features and predicted change regions through a supervised interface that exposes entity presence, temporal appearance variation, and spatial extent. Retaining dense visual memory allows the decoder to combine these explicit conditions with detailed scene evidence.  Results on LEVIR-MCI and WHU-CDC demonstrate strong joint performance and support the value of exposing visual attributes to the caption decoder. Entity semantics follow the detection ontology, the temporal target is a weak photometric proxy, and extent measures area coverage rather than individual object change. Future work can extend this interface to object level event representations and evaluate the grounding of individual generated claims, building on ESA's central principle of exposing and reusing supervised visual evidence within language generation.


{
\bibliographystyle{ieeenat_fullname}
\bibliography{main}
}

\end{document}